\documentclass[11pt,a4paper]{article}
\usepackage{times,latexsym}
\usepackage{url}
\makeatletter
\g@addto@macro{\UrlBreaks}{\UrlOrds}
\makeatother
\usepackage[T1]{fontenc}
\usepackage{footmisc}

\usepackage{sidecap}
\usepackage{csquotes}
\usepackage{caption}
\usepackage{subcaption}
\usepackage{amsmath}
\usepackage{amssymb}
\usepackage{float}
\newcommand{\norm}[1]{\left\lVert#1\right\rVert}
\newcommand{\abs}[1]{\left\lvert#1\right\rvert}

\newcommand{\method}{Word2Vec with Structure Prediction}
\newcommand{\fixmethod}{Word2Vec with Structure Constraint}
\newcommand{\denoisemethod}{Word2Vec with Denoised Structure Constraint}
\newcommand{\methodabbr}{W2VPred}
\newcommand{\yao}{Dynamic Word2Vec}
\newcommand{\yaoabbr}{DW2V}
\newcommand{\fixwen}{W2VConstr}
\newcommand{\wen}{W2VPred}
\newcommand{\denoisewen}{W2VDen}
\newcommand{\wikiphil}{WikiPhil}
\newcommand{\wikifos}{Wiki\-FoS}
\newcommand{\nyt}{NYT}
\newcommand{\sg}{Skip-Gram}
\newcommand{\cbow}{CBOW}
\usepackage{graphicx}
\usepackage{booktabs}
\usepackage{bbm}
\usepackage{color}
\usepackage{colortbl}
\usepackage[dvipsnames]{xcolor}

\usepackage{multirow}

\newcommand{\gray}{\cellcolor[gray]{.8}}

\usepackage[acceptedWithA]{tacl2018v2}

\newif\ifomitsummarizingdiscussion
\omitsummarizingdiscussiontrue

\usepackage{xspace,mfirstuc,tabulary}

\newif\iftaclinstructions
\taclinstructionsfalse 
\iftaclinstructions
\renewcommand{\confidential}{}
\renewcommand{\anonsubtext}{(No author info supplied here, for consistency with
TACL-submission anonymization requirements)}
\newcommand{\instr}
\fi

\iftaclpubformat 

\else

\fi

\title{Domain-Specific Word Embeddings with Structure Prediction
}

\author{Stephanie Brandl$^{*,1,2}$ ~ David Lassner$^{*,1,3}$ ~ Anne Baillot$^{4}$ ~  Shinichi Nakajima$^{1,3,5}$ \\
  $^{1}$TU Berlin\quad $^{2}$University of Copenhagen \quad $^{3}$BIFOLD  \\
  $^{4}$Le Mans Université \quad $^{5}$RIKEN Center for AIP\\
\texttt{brandl@di.ku.dk}, \texttt{lassner@tu-berlin.de} \\
$^{*}$Authors contributed equally.\\
  }

\date{}
\begin{document}

\maketitle

\begin{abstract}

Complementary to finding good general word embeddings, an important question for representation learning is to find dynamic word embeddings, e.g., across time or domain. Current methods do not offer a way to use or predict information on structure between sub-corpora, time or domain and dynamic embeddings can only be compared after post-alignment.
We propose novel word embedding methods that provide general word representations for the whole corpus, domain-specific representations for each sub-corpus, sub-corpus structure, and embedding alignment simultaneously.
We present an empirical evaluation on New York Times articles and two English Wikipedia datasets with articles on science and philosophy. Our method, called \textit{\method~(\wen)}, provides better performance than baselines in terms of the general analogy tests, domain-specific analogy tests, and multiple specific word embedding evaluations as well as structure prediction performance when no structure is given a priori. 
As a use case in the field of Digital Humanities we demonstrate how to raise novel research questions for high literature from the German Text Archive.

\end{abstract}
\section{Introduction}
Word embeddings \cite{mikolov2013distributed, pennington2014glove} are a powerful tool for word-level representation in a vector space that captures semantic and syntactic relations between words. They have been successfully used in many applications such as text classification~\citep{joulin2016} and machine translation~\citep{mikolov2013b}.
Word embeddings highly depend on their training corpus. For example, technical terms used in scientific documents can have a different meaning in other domains, and words can change their meaning over time---``apple'' did not mean a tech company before Apple Inc.\ was founded.
On the other hand, such \emph{local} or domain-specific representations are also not independent of each other, because most words are expected to have a similar meaning across domains.

There are many situations where a given target corpus is considered to have some \emph{structure}.  For example, when analyzing news articles, one can expect that articles published in 2000 and 2001 are more similar to each other than the ones from 2000 and 2010.  When analyzing scientific articles, uses of technical terms are expected to be similar in articles on similar fields of science. 
This implies that the structure of a corpus can be a useful side information for obtaining better word representation.

Various approaches to analyse semantic shifts in text have been proposed where typically first individual static embeddings are trained and then aligned afterwards \citep[e.g.,][]{kulkarni2015statistically,hamilton-etal-2016-diachronic, kutuzov2018diachronic,tahmasebi2018survey}. As most word embeddings are invariant with respect to rotation and scaling, it is necessary to map word embeddings from different training procedures into the same vector space in order to compare them. This procedure is usually called \textit{alignment} for which orthogonal Procrustes can be applied as has been used in \cite{hamilton-etal-2016-diachronic}.

Recently, new methods to train diachronic word embeddings have been proposed where the alignment process is integrated in the training process. \citet{bamler2017dynamic} propose a Bayesian approach that extends the skip-gram model \citep{mikolov2013distributed}. \citet{rudolph2018dynamic} analyse dynamic changes in word embeddings based on exponential family embeddings. \citet{yao2018} propose \yao\ where word embeddings for each year of the New York Times corpus are trained based on individual positive point-wise information matrices and aligned simultaneously.

We argue that apart from diachronic word embeddings there is a need to train dynamic word embeddings that not only capture temporal shifts in language but for instance also semantic shifts between domains or regional differences. It is therefore important that those embeddings can be trained on small datasets.
We therefore propose two generalizations of \yao. Our first method is called \textit{\fixmethod\ (\fixwen)}, where domain-specific embeddings are learned under regularization with any kind of structure.  This method performs well when a respective graph structure is given a priori.
For more general cases where no structure information is given, we propose our second method, called \textit{\method\ (\wen)}, where domain-specific embeddings and sub-corpora structure are learned at the same time.
\wen\ simultaneously solves three central problems that arise with word embedding representations:
\begin{enumerate}
    \item Words in the sub-corpora are embedded in the same vector space, and are therefore directly comparable without post-alignment.
    \item The different representations are trained simultaneously on the whole corpus as well as on the sub-corpora, which makes embeddings for both general and domain-specific words robust, due to the information exchange between sub-corpora.
    \item The estimated graph structure can be used  for confirmatory evaluation when a reasonable prior structure is given. \wen\ together with \fixwen\ identifies the cases where the given structure is not ideal, and suggests a refined structure which leads to an improved embedding performance, we call this method \textit{\denoisemethod}.  When no structure is given, \wen\ provides insights on the structure of sub-corpora, e.g., similarity between authors or scientific domains.
\end{enumerate}
All our methods rely on static word embeddings as opposed to currently often used contextualized word embeddings. As we learn one representation per slice such as year or author, thus considering a much broader context than contextualized embeddings, we are able to find a meaningful structure between corresponding slices. Another main advantage comes from the fact that our methods do not require any pre-training and can be run on a single GPU.

We test our methods on 4 different datasets with different structures (sequences, trees and general graphs), domains (news, wikipedia, high literature) and languages (English and German). We show on numerous established evaluation methods that \fixwen\ and \wen\ significantly outperform baseline methods with regard to general as well as domain-specific embedding quality. We also show that \wen\ is able to predict the structure of a given corpus, outperforming all baselines. Additionally, we show robust heuristics to select hyperparameters based on proxy measurements in a setting where the true structure is not known. Finally, we show how \methodabbr~can be used in an explorative setting to raise novel research questions in the field of Digital Humanities. Our code is available at \href{https://github.com/stephaniebrandl/domain-word-embeddings}{\nolinkurl{github.com/stephaniebrandl/domain-word-embeddings}}.
\section{Related Work}\label{sec:related_work}
Various approaches to track, detect and quantify semantic shifts in text over time have been proposed \citep{kim2014temporal, kulkarni2015statistically, hamilton-etal-2016-diachronic, zhang2016past, marjanen2019clustering}.

This research is driven by the hypothesis that semantic shifts occur, e.g., over time \cite{bleich2016effect} and viewpoints \cite{azarbonyad2017words}, in political debates \cite{reese2009framing} or caused by cultural developments \cite{lansdall2017content}. Analysing those shifts can be crucial in political and social studies but also in literary studies as we show in Section \ref{sec:dh}.

Typically, methods first train individual static embeddings for different timestamps, and then align them afterwards \citealp[(e.g.,][]{kulkarni2015statistically,hamilton-etal-2016-diachronic,kutuzov2018diachronic,devlin2018bert,jawahar2019contextualized, hofmann2020dynamic} and a comprehensive survey by \citealp{tahmasebi2018survey}). 
Other approaches, which deal with more general structure \citep{azarbonyad2017words, gonen2020simple} and more general applications \citep{zeng2017socialized, shoemark2019room}, also rely on post-alignment of static word embeddings \citep{grave2019unsupervised}.
With the rise of larger language models such as Bidirectional Encoder Representations from Transformers (BERT) and with that contextualized embeddings, a part of the research question has shifted towards detecting language change in contextualized word embeddings \citep[e.g.,][]{jawahar2019contextualized, hofmann2020dynamic}.\\
Recent methods directly learn dynamic word embeddings in a common vector space without post-alignment: \citet{bamler2017dynamic} proposed a Bayesian probabilistic model that generalizes the skip-gram model \cite{mikolov2013distributed} to learn dynamic word embeddings that evolve over time. 
\citet{rudolph2018dynamic} analysed dynamic changes in word embeddings based on exponential family embeddings, a probabilistic framework that generalizes the concept of word embeddings to other types of data \citep{rudolph2016exponential}. 
\citet{yao2018} proposed \yao~(\yaoabbr)\ to learn individual word embeddings for each year of the New York Times dataset (1990-2016) while simultaneously aligning the embeddings in the same vector space. 
Specifically, they solve the following problem for each timepoint $t = 1, \ldots, T$ sequentially:
\begin{align}
\min_{U_t} \textstyle &  L_{\mathrm{F}}  +\tau L_{\mathrm{R}} +   \lambda  L_{\mathrm{D}}  , \quad \mbox{ where }
\label{eq:yao_objective}\\
L_{\mathrm{F}} & = \textstyle \norm{Y_t -U_t U_t^\top}^2_F,
L_{\mathrm{R}} = \textstyle \norm{U_t}^2_F,\notag\\
L_{\mathrm{D}} & =\textstyle  \norm{U_{t-1} - U_t}_F^2 + \norm{U_t-U_{t+1}}_F^2
\label{eq:yao_terms}
\end{align}
represent the losses for data fidelity, regularization,
and diachronic constraint, respectively.
$U_t \in \mathbb{R}^{V \times d}$ is the matrix consisting of 
$d$-dimensional embeddings for $V$ words in the vocabulary, and $Y_t \in \mathbb{R}^{V \times V}$ represents the positive pointwise mutual information (PPMI) matrix 
\citep{levy2014neural}.
The diachronic constraint $L_{\mathrm{D}}$ encourages alignment of the word embeddings with the parameter $\lambda$ controlling how much the embeddings are allowed to be dynamic ($\lambda=0$: no alignment and $\lambda \rightarrow \infty$: static embeddings).

\section{Methods}
By generalizing  \yaoabbr, we propose two methods, one for the case where sub-corpora structure is given as prior knowledge, and the other for the case where no structure is given a priori.
We also argue that combining both methods can improve the performance in cases where some prior information is available but not necessarily reliable.

\subsection{\fixmethod}

We reformulate the diachronic term in Eq.~\ref{eq:yao_objective}
as
\begin{align}
L_{\mathrm{D}}
& =\textstyle 
\sum_{t'=1}^T 
W_{t,t'}^{\mathrm{diac}}
\norm{U_t - U_{t'}}_F^2
\notag\\
\text{with}~W_{t,t'}^{\mathrm{diac}}& = \mathbbm{1}(\{\abs{t-t^{\prime}}=1\}),
\label{eq:Chronological}
\end{align}
where $\mathbbm{1}(\cdot)$ denotes the indicator function.
This allows us to generalize \yaoabbr\ for different neighborhood structures:
Instead of the chronological sequence \eqref{eq:Chronological}, we assume $W \in \mathbb{R}^{T \times T}$ to be an arbitrary affinity matrix representing 
the underlying semantic structure, given as prior knowledge.


Let $D \in \mathbb{R}^{T \times T}$ be the pairwise distance matrix between  embeddings such that 
\begin{align}
D_{t,t'} = \norm{U_t - U_{t'}}_F^2,
\label{eqDistanceMatrix}
\end{align}
and we impose regularization on the distance, instead of the norm of each embeddings.
This yields the following optimization problem:
\begin{align}
\min_{U_t} \textstyle &  L_{\mathrm{F}}  +\tau L_{\mathrm{RD}} +   \lambda  L_{\mathrm{S}}  , \quad \mbox{ where }
\label{eq:objective}\\
L_{\mathrm{F}} & = \textstyle \norm{Y_t -U_t U_t^\top}^2_F,
L_{\mathrm{RD}} = \textstyle \norm{D}_F,\notag\\
L_{\mathrm{S}} & =\textstyle  \textstyle 
\sum_{t'=1}^T W_{t,t'}  D_{t,t^{\prime}}.
\label{eq:WESCNei}
\end{align}
 We call this generalization of \yaoabbr\
 \emph{\fixmethod}\ (\fixwen).



\subsection{\method}
When no structure information is given, we need to estimate the similarity matrix $W$ from the data.  We define $W$ based on the similarity between embeddings.  
Specifically, we initialize (each entry of) the embeddings $\{U_t\}_{t=1}^T$ by independent uniform distribution in $[0, 1)$.
Then, in each iteration, we compute the distance matrix $D$ by Eq.\eqref{eqDistanceMatrix},
and
set 
$\widetilde{W}$ to its (entry-wise) inverse, i.e.,
\begin{align}
\!\!
\widetilde{W}_{t,t'} & \leftarrow 
\begin{cases}
\textstyle  D_{t,t^{\prime}}^{-1} & \mbox{ for } t \ne t', \\
0 & \mbox{ for } t = t'.
\end{cases} 
\label{eq:Wupdate}
\end{align}
and normalize it according to the corresponding column and row:
\begin{align}
 W_{t,t'}  & \leftarrow
 \textstyle \frac{\widetilde{W}_{t,t'}}{\sum_{t''}\widetilde{W}_{t,t''}+\sum_{t''}\widetilde{W}_{t'',t'}} .
\label{eq:WNormalize}
\end{align}

The structure loss \eqref{eq:WESCNei}
with the similarity matrix $W$ updated by Eqs.~\ref{eq:Wupdate} and \ref{eq:WNormalize}
 constrains the distances between embeddings according to the similarity structure that is at the same time estimated from the distances between embeddings.
We call this variant \emph{\method}\ (\wen).
Effectively,
$W$ serves as a weighting factor that strengthens connections between close embeddings.


\subsection{\denoisemethod}
We propose a third method that combines \fixwen~and \wen~for the scenario where \fixwen~results in poor word embeddings because the a-priori structure is not optimal. In this case, we suggest to apply \wen~and consider the resulting structure as an input for \fixwen. This procedure needs prior knowledge of the dataset and a human-in-the-loop to interpret the predicted structure by \wen~ in order to add or remove specific edges in the \textit{new} ground truth structure. In the experiment section, we will condense the predicted structure by \wen~ into a sparse, denoised ground truth structure that is meaningful. We call this method \denoisemethod~(\denoisewen).

\subsection{Optimization}
We solve the problem \eqref{eq:objective}
iteratively for each embedding $U_t$, given the other embedings $\{U_{t'}\}_{t' \ne t}$ are fixed.
We define one epoch as complete when $\{U_t\}$ has been updated for all $t$. 
We applied gradient descent with Adam \cite{kingma2014adam} with default values for the exponential decay rates given in the original paper and a learning rate of $0.1$. The learning rate has been reduced after 100 epochs to $0.05$ and after 500 epochs to $0.01$ with a total number of 1000 epochs. Both models have been implemented in PyTorch.
\wen\ updates $W$ by Eqs.~\ref{eq:Wupdate} and \ref{eq:WNormalize} after every iteration.

\section{Experiments on Benchmark Data}
We conducted four experiments starting with well-known settings and datasets and incrementally moving to new datasets with different structures. The first experiment focuses on the general embedding quality, the second one presents results on domain-specific embeddings, the third one evaluates the method's ability to predict structure and the fourth one shows the method's performance on various word similarity tasks. In the following subsections, we will first describe the data, preprocessing and then the results. Further details on implementation and hyperparameters can be found in Appendix \ref{app:implementation}.

\begin{table}[t]
\begin{tabular}{l|r|}
Category & \#Articles \\ \hline \hline 
Natural Sciences & 8536\\ \hline 
Chemistry & 19164\\
Computer Science & 11201\\
Biology & 10988\\ \hline \hline 
Engineering \& Technology & 20091\\ \hline 
Civil Engineering & 17797\\
Electrical \& Electronic Engineering & 6809\\
Mechanical Engineering & 4978\\ \hline \hline 
Social Sciences & 17347\\ \hline 
Business \& Economics & 14747\\
Law & 13265\\
Psychology & 5788\\ \hline \hline 
Humanities & 15066\\ \hline 
Literature \& Languages & 24800\\
History \& Archaeology & 16453\\
Religion \& Philosophy \& Ethics & 19356\\\hline
\end{tabular}
\vspace{-2mm}
\caption{Categories and the number of articles in the WikiFoS dataset. One cluster contains 4 categories (rows): the top one is the main category and the following 3 are subcategories. Fields joined by \& originate from 2 separate categories in Wikipedia\textsuperscript{\ref{categories}} but were joined, according to the OECD's definition.\textsuperscript{\ref{OECD}}}
\vspace{-3mm}
\label{tab:wikifos}
\end{table}

\begin{table}[t]
\centering
\begin{tabular}{l||r|}
Category & \#Articles \\ \hline \hline 
Logic & 3394\\ \hline 
Concepts in Logic & 1455\\
History of Logic & 76\\ \hline \hline 
Aesthetics & 7349\\ \hline 
Philosophers of Art & 30\\
Literary Criticism & 3826\\ \hline \hline 
Ethics & 5842\\ \hline 
Moral Philosophers & 170\\
Social Philosophy & 3816\\ \hline \hline 
Epistemology & 3218\\ \hline 
Epistemologists & 372\\
Cognition & 8504\\ \hline \hline 
Metaphysics & 1779\\ \hline 
Ontology & 796\\
Philosophy of Mind & 976\\\hline
\end{tabular}
\vspace{-2mm}
\caption{Categories and the number of articles in the WikiPhil dataset. 
One cluster contains 3 categories: the top one is the main category and the following are subcategories in Wikipedia
}\label{tab:wiki}
\vspace{-3mm}
\end{table}

\subsection{Datasets}
We evaluated our methods on the following three benchmark datasets.
\paragraph{New York Times (NYT):}
The New York Times dataset\footnote{\href{https://sites.google.com/site/zijunyaorutgers/}{\nolinkurl{sites.google.com/site/zijunyaorutgers}}} (NYT) contains headlines, lead texts and paragraphs of English news articles published online and offline between January 1990 and June 2016 with a total of 100,945 documents. We grouped the dataset by years with 1990-1998 as the train set and 1999-2016 as the test set.

\paragraph{Wikipedia Field of Science and Technology (\wikifos):}
We selected categories of the OECD’s  list of  Fields  of  Science  and  Technology\footnote{\label{OECD}\href{http://www.oecd.org/science/inno/38235147.pdf}{\nolinkurl{oecd.org/science/inno/38235147.pdf}}} and downloaded the corresponding articles from the English Wikipedia. The resulting dataset Wikipedia Field of Science and technology (\wikifos)\ contains four clusters, each of which consists of one main category and three subcategories, with 226,386 unique articles in total (see Table \ref{tab:wikifos}). The articles belonging to multiple categories\footnote{\label{categories}\href{https://en.wikipedia.org/wiki/Wikipedia:Contents/Categories}{\nolinkurl{wikipedia.org/wiki/Wikipedia:Contents/Categories}}} were randomly assigned to a single category in order to avoid similarity because of overlapping texts instead of structural similarity.
In each category,
we randomly chose $1/3$ of the articles for the train set, and the remaining $2/3$ were used as the test set. 
\paragraph{Wikipedia Philosophy (\wikiphil):}
Based on Wikipedia's definition of categories in \textit{philosophy}, we selected 5 main categories
and their 2 largest subcategories each (see Table \ref{tab:wiki}). Categories and subcategories are based on the definition given by Wikipedia. We downloaded 41,603 unique articles in total from the English Wikipedia. Similarly to WikiFoS, the articles belonging to multiple categories were randomly assigned to a single category, and the articles in each category were divided into a train set ($1/3$) and a test set ($2/3$). 
\subsection{Preprocessing}
We lemmatized all tokens, i.e., assigned their base forms with spacy\footnote{\url{https://spacy.io}} and grouped the data by years (for NYT) or categories (for \wikiphil\ and \wikifos). For each dataset, we defined one individual vocabulary where we considered the 20,000 most frequent (lemmatized) words of the entire dataset that are also within the 20,000 most frequent words in at least 3 independent slices, i.e., years or categories. This way, we filtered out \enquote{trend} words that are of significance only within a very short time period/only a few categories. The 100 most frequent words were filtered out as stop words. We set the symmetric context window (the number of words before and after a specific word considered as context for the PPMI matrix) to 5.

\setlength{\tabcolsep}{4pt}
\begin{table}[t]
\centering

\begin{tabular}{|r|l||c|c|c|}
\parbox[t]{2mm}{\multirow{2}{*}{\rotatebox[origin=c]{90}{Data}}}&&\multicolumn{3}{c|}{general analogy tests}\\\cline{3-5}
&Method&n=1&n=5&n=10\\\hline
\parbox[t]{2mm}{\multirow{6}{*}{\rotatebox[origin=c]{90}{\nyt}}}     &   GloVe     &   9.40  &  26.41  &   33.58 \\
        & Skip-Gram   &   3.62  &  16.20  &   25.61 \\
        & CBOW        &   5.58  &   19.92 &   27.60 \\
        &  DW2V       & \gray11.27   & \gray32.88    & \gray42.97\\
        & \fixwen~(our)           &  10.90   &  \gray33.01   &  \gray43.12 \\
        & \wen~(our)              &   10.28       &  31.66        &  41.88 \\\hline
\parbox[t]{2mm}{\multirow{6}{*}{\rotatebox[origin=c]{90}{\wikifos}}}    & GloVe &    6.33       &  23.74        &   32.58 \\
        & Skip-Gram   &    3.54 &   12.09 &  15.77 \\
        & CBOW        &    4.25 &   17.47 &  26.21 \\
        & \fixwen~(our)           &  \gray11.91   &  \gray45.96   &  \gray56.88 \\
        & \wen~(our)              &  \gray11.82   &  \gray45.73   &  \gray56.40 \\\cline{2-5}
        & \denoisewen~(our) & 11.61 & 46.50* & 57.08* \\
        \hline
\parbox[t]{2mm}{\multirow{6}{*}{\rotatebox[origin=c]{90}{\wikiphil}}}   & GloVe             &   2.59        &     17.45     &      24.19 \\
        & Skip-Gram   &    2.76 &   10.18 &   17.48 \\
        & CBOW        &    3.11 &    6.61 &    9.47 \\
        & \fixwen~(our)           &         0.42  &        10.37  &        15.02 \\
        & \wen~(our)              &  \gray4.37    &  \gray31.99   &  \gray41.75 \\\cline{2-5}
        & \denoisewen~(our) &\multirow{1}{*}{5.96*}   & \multirow{1}{*}{36.21*}   &  \multirow{1}{*}{46.15*} \\\hline
    \end{tabular}
\vspace{-2mm}
    \caption{General analogy test performance for our methods, \fixwen\ and \wen, and baseline methods, GloVe, \sg, \cbow~and \yaoabbr~ averaged across ten runs with different random seeds.
    The best method and the methods that are not significantly outperformed by the best is marked with a gray background, according to the Wilcoxon signed rank test for $\alpha=0.05$. W2VDen is compared against the best method from the same data set and if it is significantly better, it is marked with a star (*).}
    \label{tab:global_analogy_tests}
\vspace{-3mm}
\end{table} 

\setlength{\tabcolsep}{6pt}
\subsection{Ex1: General Embedding Performance}\label{sec:ex1}
\begin{figure*} [t]
    \begin{minipage}[c]{0.5\textwidth}
        \includegraphics[width=\textwidth]{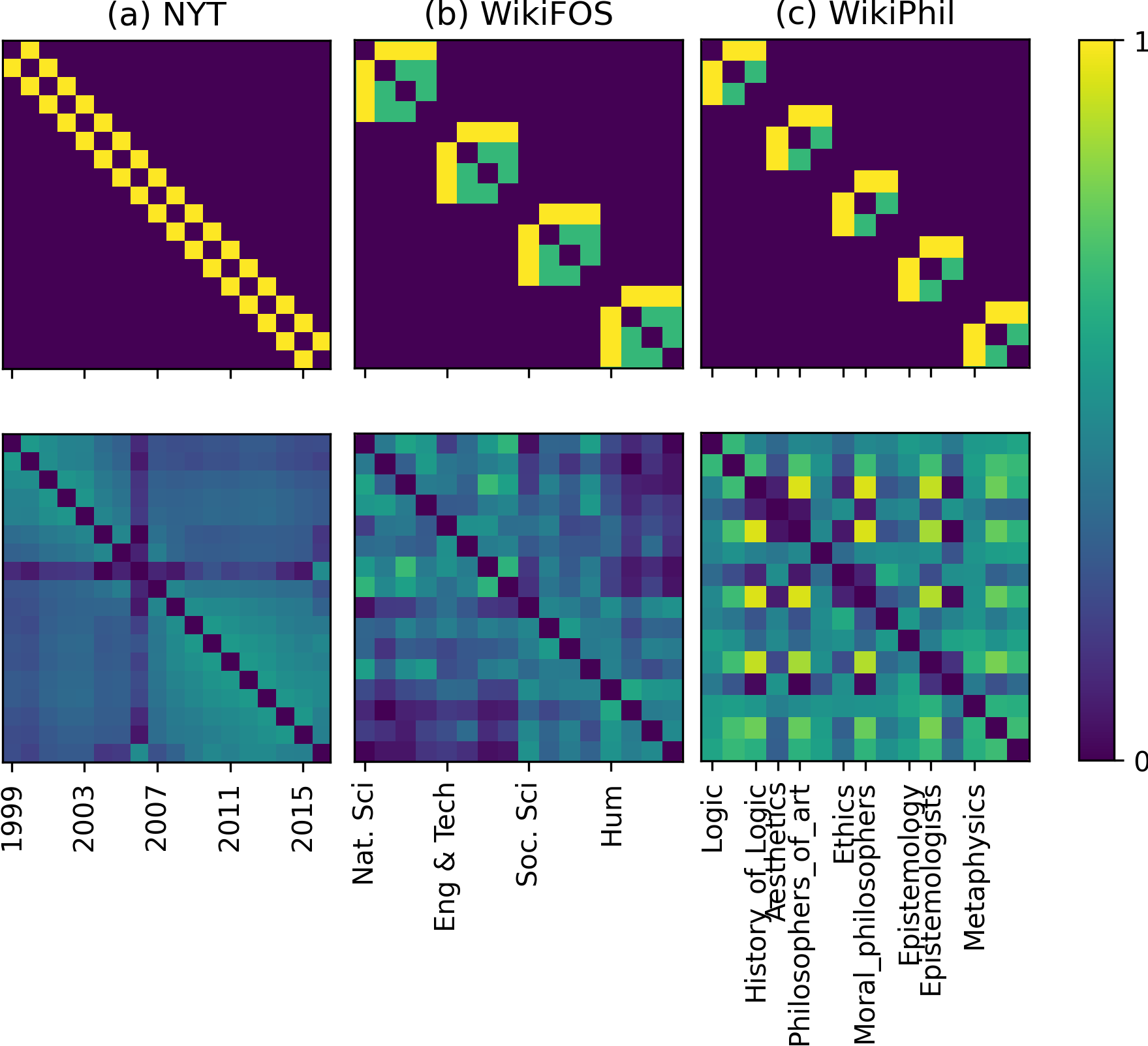}
    \end{minipage}\hfill
    \begin{minipage}[c]{0.47\textwidth}
        \caption{Prior affinity matrix $W$ used for \fixwen\ (upper), and the estimated affinity matrix by \wen\ (lower) where the number indicates how close slices are (1: identical, 0: very distant). The estimated affinity for NYT implies the year 2006 is an outlier. 
    We checked the corresponding articles and found that many paragraphs and tokens are missing in that year.  Note that the diagonal entries do not contribute to the loss for all methods.}\vspace{3cm}
    \label{fig:true_graphs}
  \end{minipage}
  
\end{figure*}

In our first experiment, we compare the quality of the word embeddings trained by \fixwen\ and \methodabbr\ with the embeddings trained by baseline methods, GloVe, Skip-Gram, CBOW and \yaoabbr. For GloVe, Skip-Gram and CBOW, we computed one set of embeddings on the entire dataset.  For \yaoabbr, \fixwen\ and \methodabbr, domain-specific embeddings $\{U_t\}$ were averaged over all domains. We use the same vocabulary for all methods.  For \fixwen, we set the affinity matrix $W$ as shown in the upper row of Figure~\ref{fig:true_graphs},  based on the a priori known structure, i.e., diachronic structure for NYT, and the category structure in Tables \ref{tab:wikifos} \& \ref{tab:wiki} for WikiFoS and WikiPhil.
The lower row of Figure~\ref{fig:true_graphs} shows the learned structure by \wen.

Specifically, we set the ground-truth affinity $W^*_{t, t'}$ as follows: for NYT, $W^*_{t,t'} = 1$ if $|t-t'| = 1$, and $W^*_{t,t'} = 0$ otherwise; for WikiFoS and WikiPhil, $W^*_{t,t'} = 1$ if $t$ is the parent category of $t'$ or vice versa, $W^*_{t,t'} = 0.5$ if $t$ and $t'$ are under the same parent category, and $W^*_{t,t'} = 0$ otherwise (see Tables~\ref{tab:wikifos} and \ref{tab:wiki} for the category structure of WikiFoS and WikiPhil, respectively, and the top row of Figure~\ref{fig:true_graphs} for the visualization of the ground-truth affinity matrices).

We evaluate the embeddings on general analogies \citep{mikolov2013distributed} to capture the general meaning of a word. Table \ref{tab:global_analogy_tests} shows the corresponding accuracies averaged across 10 runs with different random seeds.

For \nyt, \fixwen~performs similarly to DW2V, which has essentially the same constraint term---$L_{\mathrm{S}}$ in Eq.\eqref{eq:WESCNei} for \fixwen~ is the same as $L_{\mathrm{D}}$ in Eq.\eqref{eq:yao_terms} for DW2V up to scaling when $W$ is set to the prior affinity matrix for \nyt---and significantly outperforms the other baselines.  \wen~performs slightly worse then the best methods.
For \wikifos, \fixwen~and \wen~outperform all baselines by a large margin. 
In \wikiphil, \fixwen~performs poorly (worse than GloVe), while \wen~outperforms all other methods by a large margin. Standard deviation across the 10 runs are less than one for \nyt~(all methods and all $n$), slightly higher for \wikifos~and highest for \wikiphil~\methodabbr~and \fixwen~(0.28-3.17).

These different behaviors can be explained by comparing the estimated (lower row) and the a priori given (upper row) affinity matrices shown in Figure~\ref{fig:true_graphs}.
In \nyt, the estimated affinity decays smoothly as the time difference between two slices increases.  
This implies that the a priori given diachronic structure is good enough to enhance the word embedding quality (by \fixwen~and DW2V), and  estimating the affinity matrix (by \wen) slightly degrades the performance due to the increased number of unknown parameters to be estimated.
In \wikifos, although the estimated affinity matrix shows somewhat similar structure to the given one a priori, it is not as smooth as the one in \nyt~and we can recognize two instead of four clusters in the estimated affinity matrix consisting of the first two main categories (\emph{Natural Sciences} and \emph{Engineering \& Technology}), and the last two (\emph{Social Sciences} and \emph{Humanities}), which we find reasonable according to Table~\ref{tab:wikifos}.
In summary, \fixwen~and \wen~outperform baseline methods when a suitable prior structure is given.
Results on the \wikiphil\ dataset show a different tendency: the estimated affinity by \wen~is very different from the prior structure,
which implies that the corpus structure defined by Wikipedia is not suitable for learning word embeddings.  As a result, \fixwen~performs even poorer than GloVe.
Overall, Table~\ref{tab:global_analogy_tests} shows that our proposed \wen~robustly performs well on all datasets.
In Section \ref{sec:prior}, we will further improve the performance by \emph{denoising} the estimated  structure by \wen~for the case where a prior structure is not given or unreliable.

\subsection{Ex2: Domain-specific Embeddings}\label{sec:ex2}
\subsubsection{Quantitative Evaluation}
\citet{yao2018} introduced temporal analogy tests that allow us to assess the quality of word embeddings with respect to their temporal information. Unfortunately, domain-specific tests are only available for the NYT dataset. Table \ref{tab:ablation} shows temporal analogy test accuracies on the NYT dataset.
As expected, GloVe, Skip-Gram and CBOW perform poorly. We assume this is because the individual slices are too small to train reliable embeddings. The embeddings trained with \yaoabbr~and \fixwen~are learned collaboratively between slices due to the diachronic and structure terms and significantly improve the performance.
Notably, \wen\ further improves the performance by learning a more suitable structure from the data.
Indeed, the learned affinity matrix by \wen\ (see Figure~\ref{fig:true_graphs}a) suggests that not the diachronic strcuture used by \yaoabbr\ but a smoother structure is optimal.
\begin{table}[t]
\centering
\begin{tabular}{r||r|r|r|}
&n=1&n=5&n=10\\\hline
GloVe       & 7.72  & 14.39 & 17.87 \\
Skip-Gram   & 10.49 & 19.89 & 24.78 \\
CBOW        & 6.35  & 11.36 & 14.59 \\
\yaoabbr    & 39.47 & 61.94 & 67.35\\
\fixwen~(our)     & 38.23 & 57.73 & 64.54 \\
\wen~(our)        & \cellcolor[gray]{.8}41.87 & \cellcolor[gray]{.8}64.60 & \cellcolor[gray]{.8}69.67\\
\hline
\end{tabular}
\caption{Accuracies for temporal analogies (NYT).}\label{tab:ablation}
\end{table}
\begin{table*}[h]
    \centering
    \begin{tabular}{|c|c|c|c||c|c|}
    \hline
       Nat. Sci  &  Eng$\&$Tech & Soc. Sci & Hum & GloVe  & Skip-Gram\\ \hline
       generator & generator & powerful & powerful & control & Power\\
       PV & inverter & control & control & supply & inverter\\
       thermoelectric & alternator &wield & counterbalance & capacity & mover\\
       inverter & converter & drive & drive & system & electricity\\
       converter & electric & generator & supreme & internal & thermoelectric\\
       \hline
    \end{tabular}
    \vspace{-2mm}
    \caption{Five nearest neighbors to the word ``power'' in the domain-specific embedding space, learned by \wen, of four main categories of WikiFoS (left four columns),
    and in the general embedding space learned by GloVe and Skip-Gram on the entire dataset (right-most columns, respectively).
    }
    \vspace{-3mm}
    \label{tab:power}
\end{table*}
\begin{figure}[h]
    \centering
    \includegraphics[width=.5\textwidth]{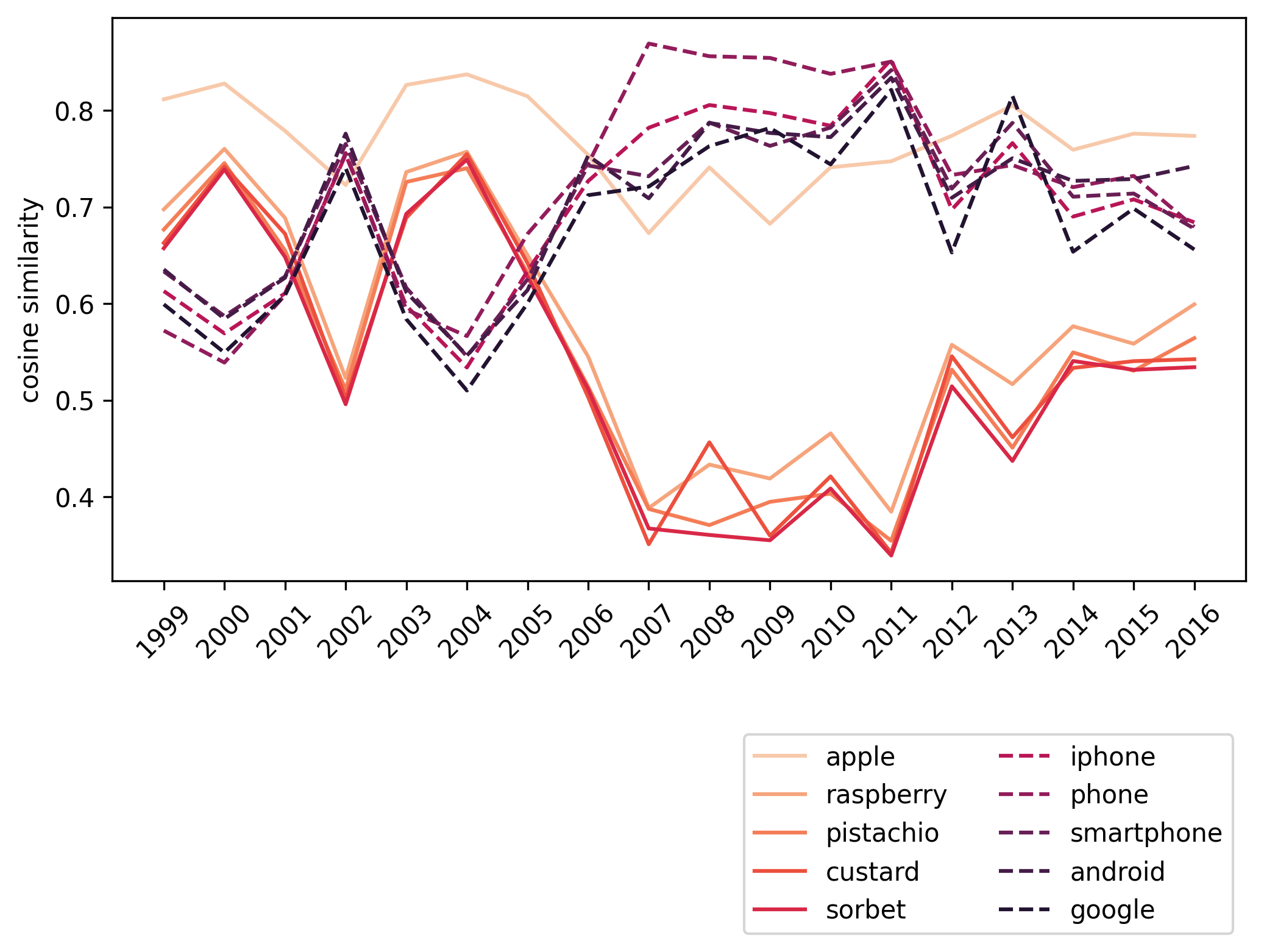}
    \caption{Evolution of the word \textit{blackberry} in NYT. Nearest neighbors of the word \textit{blackberry} have been selected in 2000 (blueish) and 2011 (reddish), and the embeddings have been computed with \wen. Cosine similarity between each neighboring word and \textit{blackberry} is plotted over time, showing the shift in dominance between fruit and smartphone brand. The word \textit{apple} also relates to both fruit and company, and therefore stays close during the entire time period.}\label{fig:blackberry}
\end{figure}
\subsubsection{Qualitative Evaluation}
Since no domain-specific analogy test is available for WikiFoS and WikiPhil,
we qualitatively analyzed the domain-specific embeddings by checking nearest neighboring words.
Table \ref{tab:power} shows the 5 nearest neighbors of the word ``power'' in the embedded spaces for the 4 main categories of \wikifos~trained by \methodabbr~ and GloVe and Skip-Gram. We averaged the embeddings obtained by \wen\ over the subcategories in each main category.
The distance between words are measured by the cosine similarity.

We see that \wen\ correctly captured the domain-specifc meaning of ``power'':
In \textit{Natural Sciences} and \textit{Engineering $\&$ Technology} the word is used in a physical context, e.g., in combination with generators which is the closest word in both categories; In \textit{Social Sciences} and \textit{Humanities} on the other hand, the nearest words are ``powerful'' and ``control'', which, in combination, indicates that it refers to ``the ability to control something or someone''.\footnote{\url{https://www.oxfordlearnersdictionaries.com/definition/english/power_1}} The embedding trained by GloVe shows a very general meaning of power with no clear tendency towards a physical or political context, whereas Skip-Gram shows a tendency towards the physical meaning. We observed many similar examples, e.g., charge:electrical-legal, performance:quality-acting, resistance:physical-social, race:championship-ethnicity.

As another example in the \nyt~corpus, 
Figure \ref{fig:blackberry} shows 
 the evolution of the word  \textit{blackberry} which can either mean the fruit or the tech company. We selected two slices (2000 $\&$ 2012) with the largest pairwise distance for the \textit{blackberry}, and chose the top-5 neighboring words from each year.
 The figure plots the cosine similarities between  \textit{blackberry} and the neighboring words.
 The time series shows how the word \textit{blackberry} evolved from being mostly associated with the fruit towards associated with the company, and back to the fruit. This can be connected to the release of their smartphone in 2002 and the decrease in sales number after 2011.\footnote{\href{https://www.businessinsider.com/blackberry-smartphone-rise-fall-mobile-failure-innovate-2019-11}{\nolinkurl{businessinsider.com/blackberry-smartphone-rise-fall-mobile-failure-innovate-2019-11}}}\footnote{\url{businessinsider.com/blackberry-phone-sales-decline-chart-2016-9}}  Interestingly, the word \textit{apple} stays relatively close during the entire time period as its word vector also (as \textit{blackberry}) reflects both meanings, a fruit and a tech company.



\subsection{Ex3: Structure Prediction}\label{sec:ex3}
This subsection discusses the structure prediction performance by \wen.
We first evaluate the prediction performance by using the a priori affinity structure as the \emph{ground-truth} structure. The results of this experiment should be interpreted with care, because we have already seen in Section~\ref{sec:ex1} that the given a priori affinity does not necessarily reflect the similarity structure of the slices in the corpus, in particular for \wikiphil.  
We then analyze the correlation between the embedding quality and the structure prediction performance by \wen, in order to evaluate the a priori affinity as the  ground-truth in each dataset.
Finally, we apply \denoisewen~which combines the benefits of both \fixwen~and \wen~
for the case where the prior structure is not suitable.

\subsubsection{Structure Prediction Performance}

Here, we evaluate the structure prediction accuracy by \wen~with the a priori given affinity matrix $D\in \mathbb{R}^{T \times T}$ (shown in the upper row of Figure \ref{fig:true_graphs}) as the ground-truth. We report on recall@k averaged over all domains.

\begin{table} [t]
\centering
\begin{tabular}{l||r|r|r|}
\textbf{Dataset} & NYT   & WikiFos & WikiPhil \\ 
\textbf{Method}  &       &         &          \\\hline
GloVe      &           67.22 &            51.66 & \cellcolor[gray]{.8}36.67 \\
Skip-Gram  &           71.11 &            54.59 & 26.67 \\
CBOW      &           65.28 &            45.00 & 23.33 \\
\wen~(our)        &  \cellcolor[gray]{.8}81.67 &            \cellcolor[gray]{.8}62.50 & 23.33 \\
Burrows'        & 55.56 & 22.92   & 6.67\\
\hline
\end{tabular}
\vspace{-2mm}
\caption{Recall@k for structure prediction performance evaluation with the prior structure (Figure \ref{fig:true_graphs} left) used as the ground-truth.}
\vspace{-3mm}
\label{tab:structure}
\end{table}

\begin{table}[]
    \centering
    \begin{tabular}{l|c}
    Dataset        & $\rho$ \\ \hline
      \nyt      & \gray 0.58  \\ 
    \wikifos    & \gray 0.65 \\ 
    \wikiphil   & \gray -0.19 \\ 
    \wikiphil (denoised)  & -0.14
    \end{tabular}
    \caption{Pearson correlation coefficients for performance on analogy tests ($n=10$) and structure prediction evaluation (recall@k) by \wen~for the parameters applied in the grid search. Linear correlation indicates that a good word embedding quality also leads to an accurate structure prediction (and vice versa). Significant correlation coefficients ($p<0.05$) are marked in gray.}
    \label{tab:gs_correlation}
\end{table}

We compare our \wen~with Burrows' Delta \cite{burrows2002}
and other baseline methods based on the GloVe, Skip-Gram, and CBOW embeddings.
Burrows' Delta is a commonly used method in stylometrics to analyze the similarity between corpora, e.g., for identifying the authors of anonymously published documents. The baseline methods based on GloVe, Skip-Gram, and CBOW simply learn the domain-specific embeddings separately, and the distances between the slices are evaluated by Equation \ref{eqDistanceMatrix}.

Table \ref{tab:structure} shows recall@k (averaged over ten trials).  As in the analogy tests, the best methods are in gray cell according to the Wilcoxon test. 
We see that \wen\ significantly outperforms the baseline methods for \nyt~and \wikifos.
For \wikiphil, we will further analyze the affinity structure in the following section.

\subsubsection{Assessment of Prior Structure}
\label{sec:StructurePredctionAssessment}
In the following, we reevaluate the aforementioned prior affinity matrix for \wikiphil~(see Figure \ref{fig:true_graphs}). Therefore, we analyse the correlation between embedding quality and structure performance and find that a suitable ground truth affinity matrix is necessary to train good word embeddings with \fixwen.
We trained \wen\ with different parameter setting for $(\lambda, \tau)$ on the train set, and applied the global analogy tests and the structure prediction performance evaluation (with the prior structure as the ground-truth).
For $\lambda$ and $\tau$, we considered log-scaled parameters in the ranges $[2^{-2}-2^{12}]$ and $[2^4-2^{12}]$, respectively, and display correlation values on  \nyt, \wikifos, and \wikiphil~in Table \ref{tab:gs_correlation}.

In \nyt\ and \wikifos, we observe clear positive correlations between the embedding quality and the structure prediction performance, which implies that the estimated structure closer to the ground truth enhances the embedding quality. 
The Pearson correlation coefficients are $0.58$ and $0.65$, respectively (both with $p<0.05$). 

Whereas Table \ref{tab:gs_correlation} for \wikiphil~does not show a clear positive correlation.
Indeed, the Pearson correlation coefficient is even negative with $-0.19$ which implies that the prior structure for \wikiphil\ is not suitable and even harmful for the word embedding performance. This result is consistent with the bad performance of \fixwen\ on \wikiphil\ in Section~\ref{sec:ex1}.

\subsubsection{Structure Discovery by \denoisewen}\label{sec:prior}
\begin{figure}[t]
    \includegraphics[width=.5\textwidth]{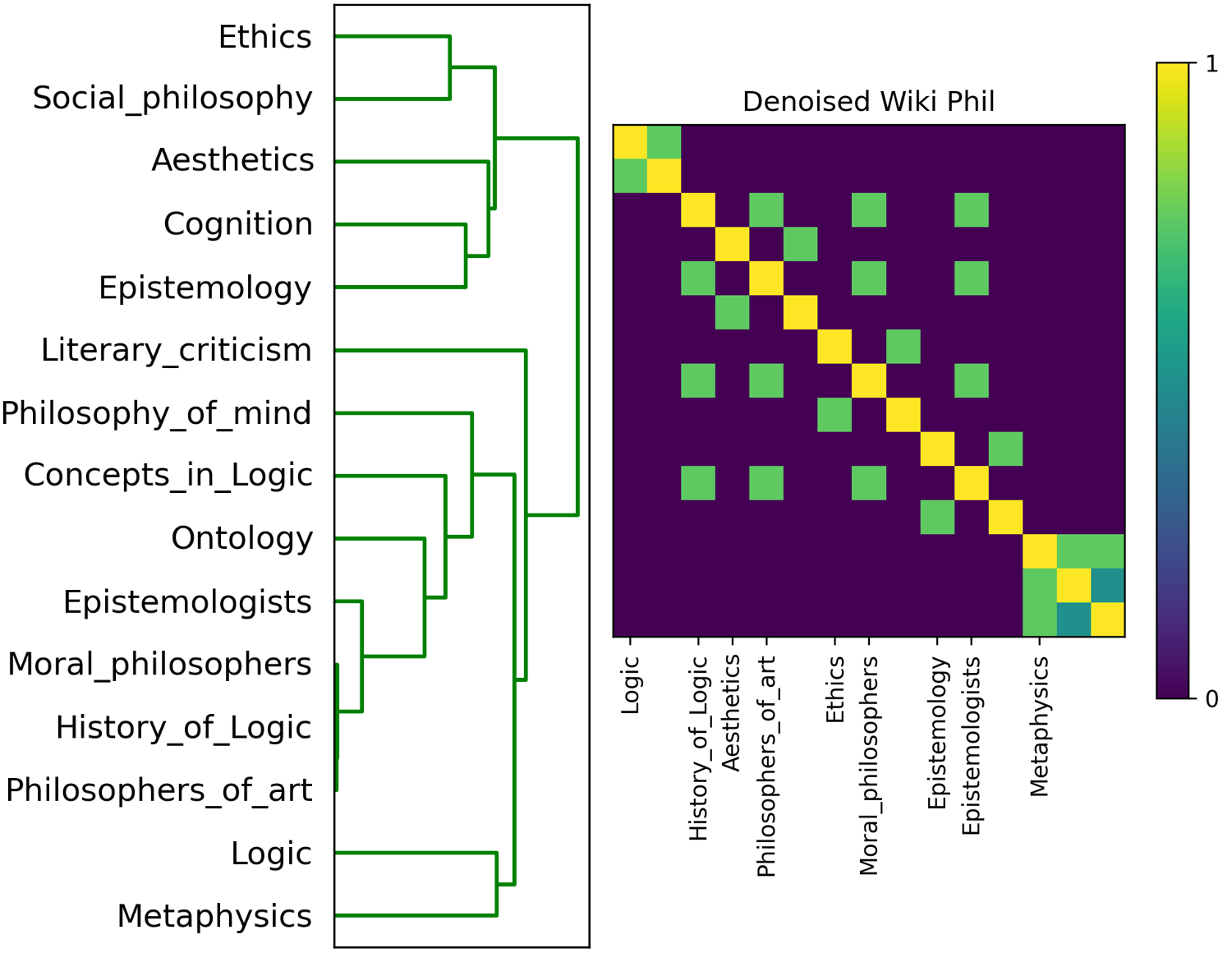}
    \vspace{-2mm}
    \caption{Left: Dendrogram for categories in \wikiphil\ learned by \wen\ based on the affinity matrix $W$. Right:Denoised Affinity matrix built from the learned structure by \wen.  Newly formed Cluster includes \textit{History of Logic}, \textit{Moral Philosophers}, \textit{Epistemologists}, and \textit{Philosophers of Art}.
    }
\vspace{-3mm}
\label{fig:denoised_graphs}
\end{figure}

The good performance of \wen\ on \wikiphil\ in Section~\ref{sec:ex1} suggests that 
\wen\ has captured a suitable structure of \wikiphil.
Here, we analyze the learned structure, and polish it with additional side information. 


Figure \ref{fig:denoised_graphs} (left) shows the dendrogram of categories in \wikiphil\ obtained from the affinity matrix $W$ learned by \wen.  
We see that  the two pairs \textit{Ethics}-\textit{Social Philosophy} and \textit{Cognition}-\textit{Epistemology} are grouped together, and both pairs also belong to the same cluster in the original structure. We also see the grouping of \textit{Epistemologists}, \textit{Moral Philosophers}, \textit{History of Logic} and \textit{Philosophers of Art}.
This was at first glance surprising because they belong to four different clusters in the prior structure.  However, investigating articles revealed that this is a natural consequence from the fact that the articles in those categories are almost exclusively about biographies of philosophers, and are therefore written in a distinctive style compared to all other slices.

To confirm that the discovered structure captures the semantic sub-corpora structure, we defined a new structure for \wikiphil, which is shown in Figure \ref{fig:denoised_graphs} (right),
based on our findings above and also define a new structure for \wikifos: A minor characteristic that we found in the structure of the prediction of \wen~in comparison with the assumed structure is that the two sub-corpora Humanities and Social Sciences and the two sub-corpora Natural Sciences and Engineering are a bit closer than other combinations of sub-corpora, which also intuitively makes sense.
We connected the two sub-corpora by connecting their root node respectively and then apply \denoisewen.
The general analogy tests performance by \denoisewen~is given in Table \ref{tab:global_analogy_tests}.
In \wikifos, the improvement is only slightly significant for $n=5$ and $n=10$ and not significant for $n=1$. This implies that the structure that we previously assumed for \wikifos~already works well. This shows that applying \denoisewen~is in fact a general purpose method that can be applied to on any of the data sets but it is especially useful when there is a mismatch between the assumed structure and the structure predicted by \wen.
In \wikiphil~, we see that \denoisewen~further improves the performance by \wen, which already outperforms all other methods with a large margin.
The correlation between the embedding quality and the structure prediction performance---with the denoised estimated affinity matrix as the ground truth---is shown in Table \ref{tab:gs_correlation}.
The Pearson correlation is still negative  $-0.14$ but not statistically significant anymore ($p=0.11$). 

\setlength{\tabcolsep}{3pt}
\begin{table*}
\begin{tabular}{l||rrrrrr@{\hspace{6pt}}|rrrrr@{\hspace{6pt}}|rrrrr|} 
& \multicolumn{6}{c}{NYT} & \multicolumn{5}{c}{WikiFos} & \multicolumn{5}{c}{WikiPhil}\\
& \rotatebox{75}{GloVe} & \rotatebox{75}{Skip-G.} & \rotatebox{75}{CBOW} & \rotatebox{75}{DW2V} & \rotatebox{75}{W2VC} & \rotatebox{75}{W2VP} & \rotatebox{75}{GloVe} & \rotatebox{75}{Skip-G.} & \rotatebox{75}{CBOW} & \rotatebox{75}{W2VC} & \rotatebox{75}{W2VP} & \rotatebox{75}{GloVe} & \rotatebox{75}{Skip-G.} & \rotatebox{75}{CBOW} & \rotatebox{75}{W2VD} & \rotatebox{75}{W2VP}  \\\hline 

RW-STANFORD & .36 & .50 & \cellcolor[gray]{.8} .55 & .51 & .52 & .53 
& .38 & \cellcolor[gray]{.8} .43 & .42 & .42 & .42 & 
.34 & \cellcolor[gray]{.8} .43 & .38 & .38 & .34 \\

MTurk-771 & .41 & .54 & .52 & .58 & \cellcolor[gray]{.8} .59 & \gray .59 
& .58 & .60 & .59 & \gray .62 & \cellcolor[gray]{.8} .62 
& .49 & .55 & .49 & \cellcolor[gray]{.8} .59 & .58 \\

RG-65 & .46 & .51 & .40 & \cellcolor[gray]{.8} .55 & .53 & \gray .54 
& .60 & .67 & .60 & \cellcolor[gray]{.8} .74 & .72 
& .42 & .50 & .38 & \cellcolor[gray]{.8} .63 & .55 \\

WS-353-ALL & .44 & .56 & .56 & \gray .57 & .56 & \cellcolor[gray]{.8} .57 
& .57 & \cellcolor[gray]{.8} .65 & .63 & .62 & .62 
& .53 & \cellcolor[gray]{.8} .59 & .57 & \gray .59 & .58 \\

MTR-3k & .50 & .63 & .60 & .68 & \cellcolor[gray]{.8} .68 & .68 
& \cellcolor[gray]{.8} .66 & .60 & .59 & .65 & .65 
& .59 & .56 & .48 & \cellcolor[gray]{.8} .62 & .60 \\

WS-353-REL & .34 & .48 & .47 & \gray .50 & .49 & \cellcolor[gray]{.8} .51 
& .52 & \cellcolor[gray]{.8} .59 & .57 & .55 & .55 .
& .51 & .54 & .51 & \cellcolor[gray]{.8} .55 & .54 \\

MC-30 & .50 & \gray .56 & .40 & \cellcolor[gray]{.8} .58 & .54 & .54 
& .71 & .68 & .59 & \cellcolor[gray]{.8} .77 & \gray .75 
& .47 & .62 & .54 & \cellcolor[gray]{.8} .66 & .58 \\


YP-13 & .38 & .37 & .31 & \gray .39 & \cellcolor[gray]{.8} .40 & \gray .40 
& .33 & .41 & .43 & \cellcolor[gray]{.8} .44 & \gray .43 
& .31 & .40 & .39 & \cellcolor[gray]{.8} .43 & .40 \\

WS-353-SIM & .53 & \gray .64 & .62 & \gray .64 & \cellcolor[gray]{.8} .64 & \gray .64 
& .63 & .67 & .66 & \cellcolor[gray]{.8} .70 & \gray .70 
& .59 & .64 & .61 & \cellcolor[gray]{.8} .67 & .64 \\

MTurk-287 & .55 & .64 & .64 & \gray .66 & \cellcolor[gray]{.8} .67 & \gray .66 
& .44 & .62 & .64 & \gray .68 & \cellcolor[gray]{.8} .69 
& .44 & .63 & .61 & \cellcolor[gray]{.8} .66 & \gray .65 \\

SimVerb-350 & .15 & .22 & \cellcolor[gray]{.8} .25 & .23 & .23 & .23 
& .17 & .17 & .20 & \gray .21 & \cellcolor[gray]{.8} .21 
& .11 & \cellcolor[gray]{.8} .19 & .17 & .19 & .18 \\

SIMLEX-999 & .26 & .32 & \cellcolor[gray]{.8} .33 & .30 & .30 & .30 
& .28 & .28 & .29 & \cellcolor[gray]{.8} .29 & \gray .29 
& .23 & \cellcolor[gray]{.8} .27 & .25 & .27 & .26 \\ \hline
Count & - & 2 & 3 & 7 & 5 & 7 & 1 & 3 & - & 8 & 7 & - & 4 & - & 9 & 1\\ \hline
\end{tabular}
\caption{Correlation values from word similarity tests on different datasets (one per row). The best method and the methods that are not significantly outperformed by the best is marked with gray background, according to the Wilcoxon signed rank test for $\alpha=0.05$. In this table, we use a shorter version of the method names (W2VC for \fixwen, etc.)}
\label{table:wordsim}
\end{table*}

\begin{table}[]
    \centering
     \begin{tabular}{l||c|c|c|}

            & NYT       & WFos      & WPhil\\ \hline
GloVe       &   0.26    &   0.29    &  0.27 \\
Skip-Gram   &   0.28    &   0.30    &  0.29 \\
CBOW        &\gray0.29  &   0.31    &  0.29 \\
\yaoabbr    &   0.28    & ---       &  --- \\
\fixwen~(our)     &\gray 0.29 & \gray 0.32& --- \\
\denoisewen~(our)     & --- & --- & \gray 0.30 \\
\wen~(our)        &   0.28    & \gray0.32 &  0.29 \\\hline
\end{tabular}
    \caption{QVEC results: correlation values of the aligned dimension between word embeddings and linguistic word vectors. }
    \label{tab:qvec}
\end{table}

\subsection{Ex4: Evaluation in word similarity tasks}
We further evaluate word embeddings on various word similarity tasks where human-annotated similarity between words is compared 
with the cosine similarity in the embedding space, as proposed in \citet{faruqui2014community}. 
Table \ref{table:wordsim} shows the correlation coefficients between the human-annotated similarity and the embedding cosine similarity,
where, again, the best method and the runner-ups (if not significantly outperformed) are highlighted.%
\footnote{We removed the dataset \emph{VERB-143} since we are using lemmatized tokens and therefore catch only a very small part of this corpus.
We acknowledge that the human annotated similarity is not domain-specific and therefore not optimal for evaluating the domain-specific embeddings.  However, we expect that this experiment provides another aspect of the embedding quality.} 
We observe that \wen~outperforms the other methods in 7 out of 12 datasets for \nyt, and \fixwen~in 8 out of 12 for \wikifos.
For \wikiphil,
since we already know that \fixwen~with the given affinity matrix does not improve the embedding performance, we instead evaluated \denoisewen, which outperforms 9 out of 12 datasets in \wikiphil.
In addition, \wen~gives comparable performance to the best method over all experiments.
 
We also apply QVEC which measures component-wise correlation between distributed word embeddings, like we use them throughout the paper, and linguistic word vectors based on WordNet \cite{fellbaum1998wordnet}. High correlation values indicate high saliency of linguistic properties and thus serve as an intrinsic evaluation method that has been shown to highly correlate with downstream task performance \cite{tsvetkov2015evaluation}. Results are shown in Table \ref{tab:qvec},
 where we observe that
 \fixwen~(as well as \denoisewen~for \wikiphil) outperforms all baseline methods, except CBOW in \nyt, on all datasets,
 and \wen~performs comparably with the best method.

\subsection{Summarizing Discussion} 
\label{sec:CombinationTwoMethods}
In this section, we have shown a good performance of \fixwen~and \wen~in terms of global and domain-specific embedding quality on news articles (\nyt) and articles from Wikipedia (\wikifos, \wikiphil). 
We have also shown that \wen\ is able to extract the underlying sub-corpora structure from \nyt\ and \wikifos.

On the WikiPhil dataset, the following observations implied that the prior sub-corpora structure, based on the Wikipedia's definition, was not suitable for analyzing semantic relations:
\begin{itemize}
\setlength\itemsep{0em}
\item Poor general analogy test performance by \fixwen~(Table~\ref{tab:global_analogy_tests}),
\item Low structure prediction performance by all methods (Table~\ref{tab:structure})
\item Negative correlation between embedding accuracy and structure score (Table \ref{tab:gs_correlation}).
\end{itemize}
Accordingly, we analyzed the learned structure by \wen, and further refined it by \emph{denoising} with human intervention.
Specifically, we analyzed the dendrogram  from Figure \ref{fig:denoised_graphs}, and found that 4 categories are grouped together that we originally assumed to belong to 4 different clusters. 
We further validated our reasoning by applying \denoisewen~with the structure shown in Figure~\ref{fig:denoised_graphs} resulting in the best embedding performance (see Table~\ref{tab:global_analogy_tests}).

This procedure poses an opportunity to obtain good global and domain-specific embeddings and extract, or validate if given a priori, the underlying sub-corpora structure by using \fixwen\ and \wen.
Namely, 
we first train \wen, and also \fixwen~if prior structure information is available.
If both methods similarly improve the embeddings in comparison with the methods without using any structure information, we acknowledge that the prior structure is at least useful for word embedding performance.
If \wen~performs well, while \fixwen\ performs poorly, we doubt that the given prior structure would be suitable, and update the learned structure by \wen.  When no prior strucuture is given, we simply apply \wen\ to learn the structure.

We can furthermore refine the learned structure with side information, which results in a clean and human interpretable structure.  Here \denoisewen~is used to validate the new structure, and to provide enhanced word embeddings. In our experiment on the WikiPhil dataset, the embeddings obtained this way significantly outperformed all other methods.
The improved performance from \wen\ is probably due to the fewer degrees of freedom of \fixwen, i.e., once we know a reasonable structure, the embeddings can be more accurately trained with the fixed affinity matrix.

\section{Application on Digital Humanities}\label{sec:dh}
\begin{figure}[t]
\includegraphics[width=.5\textwidth]{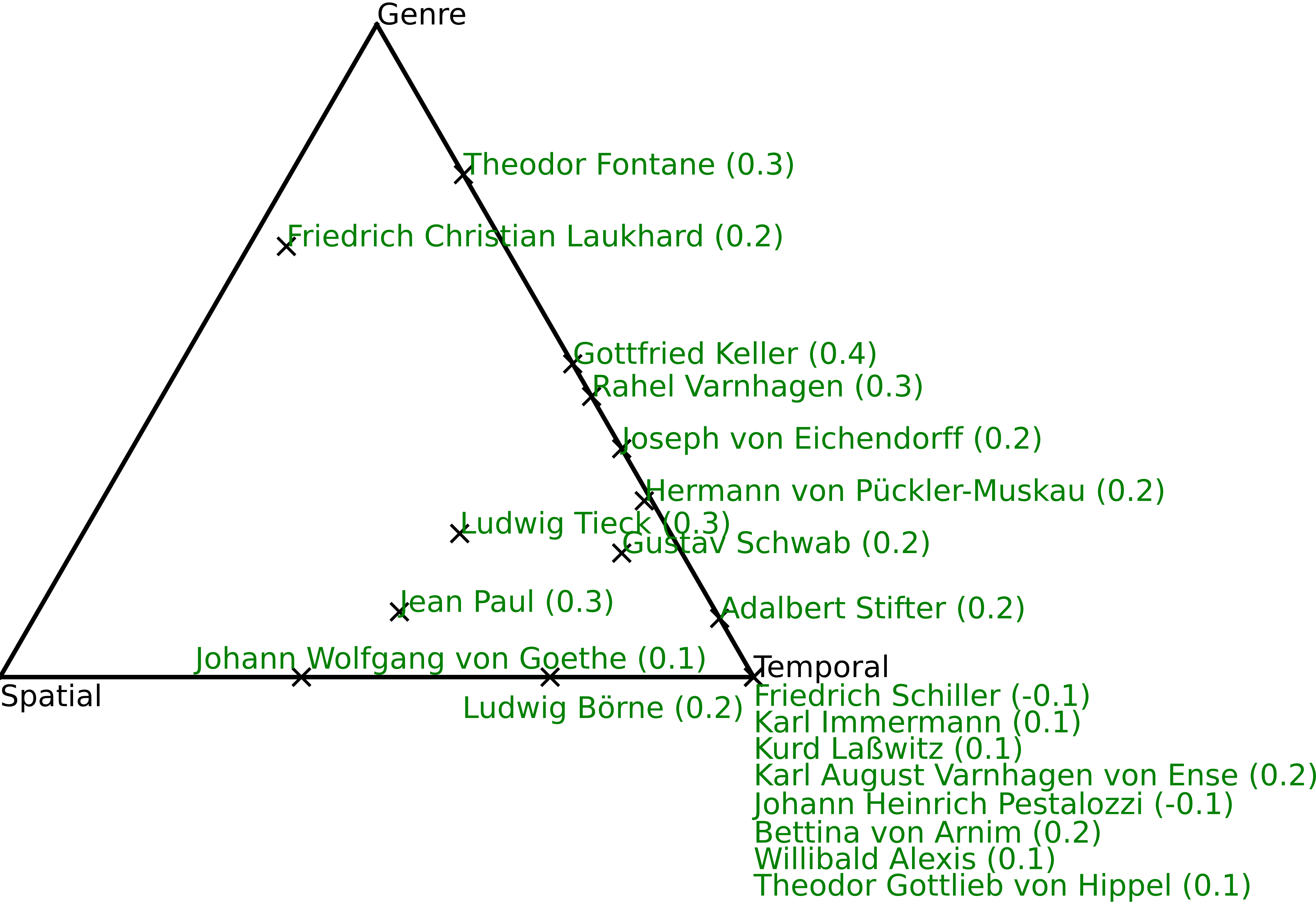}
\caption{Author's points in a barycentric coordinates triangle denote the mixture of the prior knowledge that has the highest correlation (in parentheses) with the predicted structure of \methodabbr. The correlation excludes the diagonal, meaning the correlation between the author itself.}
\label{fig:dtacorrelationwithexclusion}
\vspace{-5mm}
\end{figure}
We propose an application of \methodabbr~to the field of Digital Humanities, and develop an example more specifically related to Computational Literary Studies. In the renewal of literary studies brought by the development and implementation of computational methods, questions of authorship attribution and genre attribution are key to formulating a structured critique of the classical design of literary history, and of Cultural Heritage approaches at large. In particular the investigation of historical person networks, knowledge distribution and intellectual circles has shown to benefit significantly from computational methods \cite{baillot2018, moretti2005graphs}. Hence, our method and its capability to reveal connections between sub-corpora (such as authors' works), can be applied with success to these types of research questions.
Here, the use of quantitative and statistical models can lead to new, hitherto unfathomed insights. A corpus-based statistical approach to literature also entails a form of emancipation from literary history in that it makes it possible to shift perspectives, e.g., to reconsider established author-based or genre-based approaches.

To this end, we applied \methodabbr~to high literature texts (\textit{Belletristik}) from the lemmatized versions of DTA (German Text Archive), a corpus selection that contains the 20 most represented authors of the DTA text collection for the period 1770-1900. We applied \methodabbr~in order to predict the connections between those authors with $\lambda=512, \tau=1024$ (same as \wikifos).

As a measure of comparison, we extracted the year of publication as established by DTA, and identified the place of work for each author\footnote{via the German Integrated Authority Files Service (GND) where available, adding missing data points manually} and categorized each publication into one of three genre categories (ego document, verse and fiction).
Ego documents are texts written in the first person that document personal experience in their historical context. They include letters, diaries, memoirs and have gained momentum as a primary source in historical research and literary studies over the past decades.
We created pairwise distance matrices for all authors based on the spatial, temporal and genre information. Temporal distance was defined as the absolute distance between the average publication year, the spatial distance as the geodesic distance between the average coordinates of the work places for each author and the genre difference as cosine distance between the genre proportions for each author.
For each author, we correlated linear combinations of this (normalized) spatio-temporal-genre prior knowledge with the structure found by our method which we show in Figure \ref{fig:dtacorrelationwithexclusion}.

\paragraph{Reference Dimensions}
In this visualization we want to compare the pairwise distance matrix that our method predicted with the distance matrices that can be obtained by meta data available in the DTA corpus - the reference dimensions: 
\begin{enumerate}
	\item Temporal difference between authors. We collect the publication year for each title in the corpus and compute the average publication year for each author. The temporal distance between one author $A_{t1}$ and another author $A_{t2}$ is computed by $|A_{t1}-A_{t2}|$, the absolute difference of the average publication year.
	\item Spatial difference between authors. We query the German Integrated Authority File for the authors' different work places and extract them as longitude and latitude coordinates on the earths surface. We compute the average coordinates for each author by converting the coordinates into cartesian system and take the average on each dimension. Then, we convert the averages back into the latitude, longitude system. The spatial distance between two authors is computed by the geodesic distance as implemented in geopy.\footnote{\url{https://geopy.readthedocs.io/en/stable/}}
	\item Genre difference between authors. We manually categorized each title in the corpus into one of the three categories ego document, verse and fiction. A genre representation for an author $A_g=(A_{g_\text{ego}}, A_{g_\text{verse}}, A_{g_\text{fiction}})$ is the relative frequency of the respective genre for that author. The distance between one author $A_{g1}$ and another author $A_{g2}$ is computed by $1 - \frac{A_{g1}\cdot A_{g2}}{||A_{g1}||\cdot||A_{g2}||}$ , the cosine distance.
\end{enumerate}
\paragraph{Calculating the Correlations}
For each author $t$, we denote the predicted distance to all other authors as $X_t\in\mathbb{R}^{T-1}$ where $T$ is the number of all authors. $Y_t\in\mathbb{R}^{(T-1)\times 3}$ denotes the distances from the author $t$ to all other authors in the three meta data dimensions: space, time and genre. For the visualization we seek for the coefficients of the linear combination of $Y$ that has the highest correlation with $X$. For this, Non-Negative Canonical Correlation Analysis with one component is applied. The MIFSR algorithm is used as described by \citet{sigg2007}.\footnote{We use $\epsilon=.00001$} The coefficients are normalized to comply with the sum-to-one constraint for projection on the 2d simplex.\\

For many authors, the strongest correlation occurs with a mostly temporal structure and fewer correlate strongest with the spatial or the genre model.
Börne and Laukhard who have a similar spatial weight and thereby forming a spatial cluster, both resided in France at that time. The impact of French literature and culture on Laukhard and Börne's writing deserves attention, as suggested by our findings.

For Fontane, we do not observe a notable spatial proportion which is surprising because his sub-corpus mostly consists of ego documents describing the history and geography of the area surrounding Berlin, his workplace. 
However, in contrast to the other authors residing in Berlin, the style is a lot more similar to a travel story. In \methodabbr 's predicted structure, the closest neighbor of Fontane is, in fact, Pückler (with a distance of .052), who also wrote travel stories.

In the case of Goethe, 
the maximum correlation at the (solely spatio-temporal) resulting point is relatively low and interestingly, the highest disagreement between \methodabbr~and the prior knowledge is between Schiller and Goethe. The spatio-temporal model represents a close proximity; however, in \methodabbr's found structure, the two authors are much more distant. In this case, the spatio-temporal properties are not sufficient to fully characterize an author's writing and the genre distribution may be skewed due to the incomplete selection of works in the DTA and due to the limitations of the labelling scheme, as in the context of the 19th century, it is often difficult to distinguish between ego documents and fiction.

Nonetheless we want to stress the importance of the analysis where linguistic representation and structure, captured in \methodabbr , is in line with these properties and also, where they disagree. Both agreement and disagreement between the prior knowledge and the linguistic representation found by \methodabbr~can help identifying the appropriate ansatz for a literary analysis of an author.

\section{Conclusion}
We proposed novel methods to capture domain-specific semantics, which is essential in many natural language processing (NLP) tasks:
 \fixmethod~(\fixwen) trains domain-specific word embeddings based on prior information on the affinity structure between sub-corpora; \method~(\wen) goes one step further and predicts the structure while learning domain-specific embeddings simultaneously. Both methods outperform baseline methods in benchmark experiments with respect to embedding quality and the structure prediction performance. Specifically, we showed that embeddings provided by our methods are superior in terms of global and domain-specific analogy tests, word similarity tasks, and the QVEC evaluation, which is known to highly correlate with downstream performance.
The predicted structure is more accurate than the baseline methods including Burrows' Delta. We also proposed and successfully demonstrated a procedure, \denoisemethod~(\denoisewen),  to cope with the case where the prior structure information is not suitable for enhancing embeddings, by using both \fixwen\ and \wen.
Overall, we showed the benefits of our methods, regardless of whether (reliable) structure information is given or not.
Finally, we were able to demonstrate how to use \methodabbr~to gain insight into the relation between 19th century authors from the German Text Archive and also how to raise further research questions for high literature.

\section*{Acknowledgements}
SB was partially funded by the Platform Intelligence in News project, which is supported by Innovation Fund Denmark via the Grand Solutions program and from the European
Union’s Horizon 2020 research and innovation programme under the Marie Skłodowska-Curie grant
agreement No.~101065558. DL and SN are supported by the German Ministry for Education and Research (BMBF) as BIFOLD - Berlin Institute for the Foundations of Learning and Data under grants 01IS18025A and 01IS18037A. We thank Gilles Blanchard for valuable comments on the manuscript.

\bibliography{bibliography}
\bibliographystyle{acl_natbib}

\appendix

\section{Implementation Details}\label{app:implementation}
\subsection{Ex1}\label{app:ex1}
All word embeddings were trained with $d=50$.
\paragraph{GloVe}
We run GloVe experiments with $\alpha=100$ and minimum occurrence~$ = 25$.
\paragraph{Skip-Gram, CBOW}
We use the Gensim \cite{rehurek_lrec} implementation of Skip-Gram and CBOW with min\_alpha $= 0.0001$, sample $= 0.001$ to reduce frequent words and for Skip-Gram, we use 5 negative words and ns\_component $= 0.75$.
\paragraph{Parameter selection}
The parameters $\lambda$ and $\tau$ for \yaoabbr, \fixwen\ and \methodabbr~were selected based on the performance in the analogy tests on the train set. In order to flatten the contributions from the $n$ nearest neighbors (for $n=1, 5, 10$), we rescaled the accuracies:  For each $n$, accuracies are scaled so that the best and the worst method is 1 and 0, respectively.  Then, we computed their average and maximum.
\paragraph{Analogies}
Each analogy consists of two word pairs (e.g., countryA - capitalA; countryB - capitalB). We estimate the vector for the last word by $\widehat{v} = $ capitalA - countryA + countryB, and check if capitalB is contained in the $n$ nearest neighbors of the resulting vector $\widehat{v}$.
\subsection{Ex2}
\paragraph{Temporal Analogies}
Each of two word pairs consists of a year and a corresponding term, as e.g., 2000 - Bush; 2008 - Obama,
and the inference accuracy of the last word by vector operations on the former three tokens in the embedded space is evaluated.
To apply these analogies, GloVe, Skip-Gram and CBOW are trained individually on each year on the same vocabulary as \wen~(same parameters for GloVe as before, with minimum occurrence=10).
For the other methods, \yaoabbr, \fixwen, and \wen,
we can simply use the embedding obtained in Section~\ref{sec:ex1}.
Note that the parameters $\tau$ and $\lambda$ were optimized based on the general analogy tests.

\subsection{Ex3}

\paragraph{Burrows} It compares normalized bag-of-words features of documents and sub-corpora, and provides a distance measure between them. Its parameters specify which word frequencies are taken into account. We found that considering the 100th to the 300th most frequent words gives the best structure prediction performance on the train set. 

\paragraph{Recall@k} Let $\hat D\in \mathbb{R}^{T \times T}$ be the predicted structure.  We report on recall@k averaged over all domains:
\begin{align}
    \text{recall@k} &= \textstyle \frac{1}{T}\sum_t^T \text{recall@k}_t,
    \qquad \mbox{where}
    \notag
    \end{align}
$
    \text{recall@k}_t = \frac{\text{TP}_t(k)}{\text{TP}_t(k) + \text{FN}_t(k)}$,\\
$\text{TP}_t(k) = \sum_{t^\prime}^T b(D_t, t^\prime, k) ~\&~ b(\hat D_t, t^\prime, k)$, 
    $\text{FN}_t(k) = \sum_{t^\prime}^T b(D_t, t^\prime, k) ~\&~ \neg b(\hat D_t, t^\prime, k)$, and 
\begin{align}
    b(x, i, k) = \textstyle
    \begin{cases} 
      1 & x_i\text{ is one of the k smallest in } x, \\
      0 & \text{otherwise}.
  \end{cases}
  \notag
\end{align} 
For \nyt, we chose $k=2$, which means relevant nodes are the two next neighbors, i.e., the preceding and the following years. For \wikifos\ and \wikiphil, we respectively chose $k=3$ and $k=2$, which corresponds to the number of subcategories that each main category consists of.

\paragraph{\wen} Hyperparameters for \wen~were selected on the train set where we maximized the accuracy on the global analogy test as before.


\end{document}